%% file: Main_revised.tex
\documentclass[letterpaper, 10 pt, journal, twoside]{IEEEtran}

\usepackage{cite}
\usepackage{float}
\usepackage{multirow}
\usepackage{amssymb,amsfonts}
\usepackage{algorithmic}
\usepackage{booktabs}
\usepackage{graphicx}
\usepackage{textcomp}
\usepackage[dvipsnames]{xcolor}
\usepackage{algorithm}
\usepackage{verbatim} 
\usepackage{mathrsfs}
\usepackage[caption=false,font=footnotesize]{subfig}
\usepackage{empheq}
\usepackage{lipsum}
\usepackage{comment}
\usepackage{mathtools, cuted}
\usepackage{bm}

\usepackage{amsthm}
\usepackage[utf8]{inputenc}
\usepackage[english]{babel}
\usepackage[colorlinks,linkcolor=black,citecolor=black,urlcolor=black,bookmarks=false,hypertexnames=true]{hyperref} 
\usepackage[normalem]{ulem}
\usepackage[textsize=tiny]{todonotes}
\usepackage[font=footnotesize,skip=4pt]{caption}
\usepackage{etoolbox}
\apptocmd{\thebibliography}{\scriptsize}{}{}

\IEEEoverridecommandlockouts
\usepackage{soul}

\newtheorem*{definition*}{Definition}



\begin{document}
\title{HINT-VLM: Humanoid Interactive Navigation over Rough Terrain with Embodiment-Grounded Vision-Language Model}

% Title
% Robot-grounded vs. Embodiment-grounded vs. Embodiment-aware vs. Embodied vs. Embodiment-specific
% ?: in the results. no question mark vs. no QnA format vs. 
% Let's upload it to Arxiv before CoRL 
% next step: long-range, campus wide-level navigation + denser objects + better interaction with VLA

% Arxiv this ->  T-FR (treat Arxiv as a separate one) on Jan -> then this version to IROS on Mar?

% \author{TBD$%
% \author{Anonymous Author(s)%
\author{Ziwon Yoon$^{1}$, Seok Hwan Kim$^{1}$, Yiyu Chen$^{2}$, Quan Nguyen$^{3}$, Lu Gan$^{1}$, and Ye Zhao$^{1}$%
\thanks{$^{1}$Institute for Robotics and Intelligent Machines, Georgia Institute of Technology, Atlanta, GA 30332, USA
        {\tt\footnotesize \{zyoon6, skim3307, lgan, yezhao\}@gatech.edu}}%
\thanks{$^{2}$VinMotion
        {\tt\footnotesize v.cheny@vinmotion.net}}%
\thanks{$^{3}$Department of Aerospace and Mechanical Engineering, University of Southern California, Los Angeles, CA 90089, USA
        {\tt\footnotesize quann@usc.edu}}%
\thanks{This work was in part supported by VinMotion.}%
% \thanks{$^{1}$Institute for Robotics and Intelligent Machines, Georgia Institute of Technology, Atlanta, GA 30332, USA 
%         {\tt\footnotesize \{zyoon6 lgan, yezhao\}@gatech.edu}
%         }%

% \thanks{Digital Object Identifier (DOI): see top of this page.}
}

%%%%%%%%%% Leave here commented out for PaperCept submission %%%%%%%%%%

% \markboth{TBA}
% {Anonymous Author \MakeLowercase{\textit{et al.}}: Humanoid Interactive navigation planning in cluttered rough terrain with Vision Language Model} 

%%%%%%%%%% Leave here commented out for PaperCept submission %%%%%%%%%%

% \IEEEaftertitletext{\vspace{-6\baselineskip}}
\maketitle

%%%%%%%%%% Leave this for PaperCept submission %%%%%%%%%%
%% The following lines are needed for disabling headers
\thispagestyle{empty}
\pagestyle{empty}
%%%%%%%%%% Leave this for PaperCept submission %%%%%%%%%%

%%%%%%%%%%%% Auxiliary: For comments %%%%%%%%%%%%
\newcommand{\lu}[1]{\textbf{\textcolor{Emerald}{[Lu: #1]}}}
\newcommand{\ye}[1]{\textbf{\textcolor{brickRed}{Ye: #1}}}
\newcommand{\ziwon}[1]{\textbf{\textcolor{blue}{#1}}}
\newcommand{\Ye}[1]{\textbf{\textcolor{red}{Ye: #1}}}
\newcommand{\removed}[1]{\sout{#1}}
\newcommand{\insertfig}{\includegraphics[width=\linewidth, height=100pt]{example-image}}% define the image
% Keep draft builds usable while clearly marking figure assets that have not been added yet.
\newcommand{\includegraphicsfallback}[2][]{%
  \IfFileExists{#2}{\includegraphics[#1]{#2}}{%
    \fbox{\parbox[c][0.16\textheight][c]{0.95\linewidth}{\centering Missing figure:\\[2pt]\texttt{\detokenize{#2}}}}%
  }%
}

%%%%%%%%%%%%%%%%%%%%%%%%%%%%%%%%%%%%%%%%%%%%%%%%%%%%%%%%%%%%%%%%%%%%%%%%%%%%%%%%

\begin{abstract}
Humanoid robots have achieved significant progress in locomotion and manipulation, enabling traversal over rough terrain and richer interaction with the environment, such as moving obstacles. However, most traditional navigation methods assume non-interactive environments and focus only on traversal. While Navigation Among Movable Obstacles (NAMO) approaches relax this assumption, prior efforts have largely focused on flat environments and reactive interaction, rather than predicting object movability and planning ahead. These limitations become critical in outdoor rough-terrain scenarios, where robots must jointly reason about terrain traversability and object movability. In this work, we address NAMO for humanoid robots in cluttered outdoor rough terrain by proposing a hybrid planning framework that combines semantic priors from a Vision-Language Model (VLM) with embodied robot interaction data through a Gaussian process (GP) residual model. The GP corrects the VLM-derived movability prior to produce embodiment-grounded probabilistic movability estimates, which are then incorporated into hierarchical uncertainty-aware planning. The proposed approach allows humanoids to jointly leverage locomotion and manipulation for efficient and safe navigation, representing the first solution to NAMO in rough terrain through predictive movability reasoning and probabilistic VLM integration. Project page: \href{https://yzwfromk.github.io/HINT_VLM/}
{\textcolor{blue}{\nolinkurl{https://yzwfromk.github.io/HINT_VLM/}}} % \Ye{change the domain name to be the one only related to the work method. Remove anonymous.4open.science. - Ziwon: If you are talking about the true URL, I asked GPT and it says no way to customize the name. For State-nav, I created a new github account but cannot now.}
\end{abstract}

% \begin{IEEEkeywords}
% \end{IEEEkeywords}
\vspace{-0.1in}

\input{Introduction.tex}
\input{Related_Works.tex}
\input{VLM.tex}
\input{GSAM_TerrainRecon.tex}
\input{Path_Planning.tex}
\input{Experimental_Results.tex}
\input{Conclusion_and_Future_Works.tex}

%%%%%%%%%%%%%%%%%%%%%%%%%%%%%%%%%%%%%%%%%%%%%%%%%%%%%%%%%%%%%%%%%%%%%%%%%%%%%%%%

%%%%%%%%%%%%%%%%%%%%%%%%%%%%%%%%%%%%%%%%%%%%%%%%%%%%%%%%%%%%%%%%%%%%%%%%%%%%%%%%

%%%%%%%%%%%%%%%%%%%%%%%%%%%%%%%%%%%%%%%%%%%%%%%%%%%%%%%%%%%%%%%%%%%%%%%%%%%%%%%%
% \section*{APPENDIX}

% \subsection{Equimomental Ellipsoids}
% Add partial derivatives computed by MATLAB if we have enough space.

% \subsection{Linear Time Varying MPC Dense QP Formulation}

%\section*{ACKNOWLEDGMENT}

%The preferred spelling of the word ÒacknowledgmentÓ in America is without an ÒeÓ after the ÒgÓ. Avoid the stilted expression, ÒOne of us (R. B. G.) thanks . . .Ó  Instead, try ÒR. B. G. thanksÓ. Put sponsor acknowledgments in the unnumbered footnote on the first page.

%%%%%%%%%%%%%%%%%%%%%%%%%%%%%%%%%%%%%%%%%%%%%%%%%%%%%%%%%%%%%%%%%%%%%%%%%%%%%%%%

%put external references inside 

\let\secfnt\undefined
\newfont{\secfnt}{ptmb8t at 10pt}

\bibliographystyle{IEEEtran}
\bibliography{bibliography/external_ref}

\end{document}

%% file: Introduction.tex
\section{Introduction}\label{sec:introduction}

%\textcolor{red}{[the intro is too long. We need to cut at least one-quarter page to save space.]}

\IEEEPARstart{H}{umanoid} robots have drawn significant interest due to their ability to traverse diverse rough terrains and actively interact with the environment through whole-body manipulation using their hands or limbs~\cite{gu2026humanoid}. In parallel with this growing interest, both locomotion and manipulation capabilities have seen substantial progress~\cite{he2025attention, wei2026psi_0, liu2025opt2skill}. However, comparatively little attention has been given to navigation problems that require these two capabilities to be used jointly in cluttered rough environments. %\removed{In particular, the role of whole-body manipulation in enabling active navigation through environmental interaction remains underexplored, whereas navigation over static rough terrain has been studied in several works~\cite{yoon2025state}.} 

\begin{figure}[t]
\centerline{\includegraphicsfallback[width=1.0\linewidth]{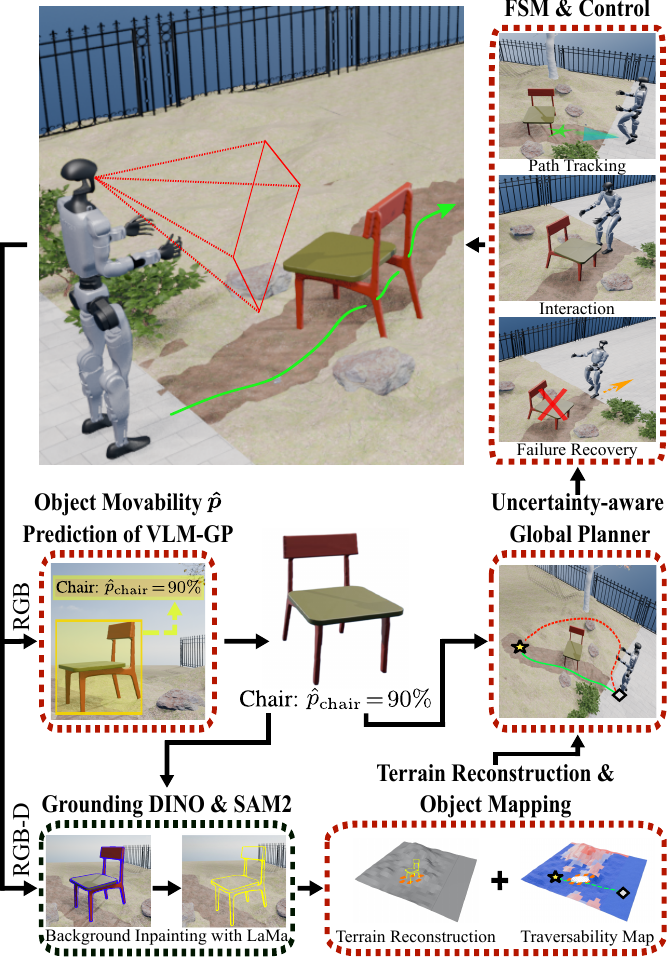}}
\caption{Overview of HINT-VLM. The proposed framework uses an embodiment-grounded VLM-GP to predict object movability from visual observations, maps objects and reconstructs the underlying terrain, and incorporates the resulting information into an uncertainty-aware global planner. A finite-state machine switches among path tracking, interaction, and failure recovery.}
\label{fig:overview}
  \vspace{-1.0\baselineskip}
\end{figure}

Most existing navigation approaches treat obstacles as entities to avoid rather than objects that the robot can physically interact with to facilitate navigation~\cite{yoon2025state, castro2023does}. More recently, a separate line of work studies Navigation Among Movable Obstacles (NAMO), or Interactive Navigation, where robots proactively interact with and manipulate objects to reach a goal~\cite{ramesh2025probe, schoch2024sight}. Existing NAMO approaches~\cite{ramesh2025probe} have primarily focused on wheeled or quadrupedal platforms, where interaction is often limited to simple forward pushing with the body in relatively flat indoor environments. This setting presents a natural opportunity for humanoid robots, which combine bipedal mobility over rough terrain with object manipulation capability.  %capabilities and navigate over more generalized terrains through bipedal locomotion. 
While recent work has begun to explore wheeled humanoids with bimanual pushing, the range of object geometries remains limited and the evaluation is still restricted to flat terrain~\cite{dai2025interactive}. Consequently, handling a wide variety of objects within NAMO, particularly in rough terrain settings, remains largely unexplored.

Another limitation of existing NAMO research lies in its predictive planning capability. Most prior works~\cite{ramesh2025probe, dai2025interactive} adopt reactive strategies that focus on improving manipulation performance or detecting failures after the physical interaction has already been initiated, often resulting in trial-and-error object interactions. In contrast, relatively few studies~\cite{schoch2024sight, wang2020affordance} address the problem of identifying movable obstacles at a distance and selecting interaction targets before physical contact, which is critical for efficient long-horizon planning. %\removed{This gap is not surprising, as inferring interaction feasibility directly from visual observations using conventional neural networks remains challenging and typically requires large amounts of task-specific data.} %\textcolor{red}{[what gap? Did not mention before. Can we remove this sentence? if you think it is important, then build a bit more context for the last sentence.]} 

The need for predictive planning becomes more pronounced in rough-terrain settings, where the robot must reason jointly about path length, terrain traversability, and object movability. In such scenarios, successful displacement depends not only on the object itself but also on the terrain conditions beneath, which influence the outcome of interaction. As a result, a humanoid robot must jointly plan rough-terrain traversal and object interactions based on its locomotion and manipulation capabilities to determine an optimal navigation path. This setting motivates the study of NAMO in outdoor, unstructured environments, which remains largely unexplored.

Fortunately, recent advances in Vision-Language Models (VLMs) offer a promising approach for inferring object movability before physical contact. Through large-scale pretraining, VLMs acquire strong language-conditioned semantic understanding and generalization capabilities that are difficult to achieve with conventional neural networks trained from scratch for language-conditioned navigation tasks~\cite{cheng2024navila, castro2025vamos, song2025vl, oh2025language}. Motivated by these capabilities, we propose to leverage VLMs to infer object movability without requiring large-scale, language-annotated training data. This enables predictive planning by allowing the robot to determine which obstacles are preferable to manipulate before committing to physical interaction. However, directly treating VLM predictions as reliable estimates can be problematic because their generic semantic priors do not reflect the physical capabilities of a specific robot embodiment and controller. We therefore correct the VLM prior using a residual Gaussian process (GP) learned from robot interaction data, producing embodiment-grounded movability estimates. Movability is represented probabilistically and incorporated into uncertainty-aware planning, allowing the planner to explicitly account for the potential cost of failed object interactions.

In this work, we address \underline{h}umanoid \underline{i}nteractive \underline{n}avigation on rough \underline{t}errain (HINT) by integrating embodiment-grounded VLM-GP-based object movability reasoning with an uncertainty-aware planner. %\removed{proposing a navigation system that integrates a hierarchical uncertainty-aware planning framework with semantic reasoning from VLM.}\textcolor{red}{[semantic reasoning for object or terrain?]} 
As illustrated in Fig.~\ref{fig:overview},
the proposed framework integrates vision-based object movability reasoning, object
and terrain mapping, uncertainty-aware global planning, and interaction
execution within a unified navigation system. The key contributions of this paper are summarized as follows:
\begin{itemize}
    \item \textbf{NAMO in cluttered rough terrain with humanoids:} To the best of our knowledge, this is the first work to address the NAMO problem in cluttered rough terrain for humanoids, jointly considering terrain traversability and object interaction, while enabling dynamic humanoid  navigation with object manipulation.
    
    \item \textbf{Predictive object movability reasoning using an embodiment-grounded VLM-GP:} We use a VLM-GP to estimate object movability as the probability of successful object interaction from visual observations, enabling the robot to reason about obstacle movability before physical interaction. The VLM prior is refined by a residual GP learned from robot interaction outcomes to produce robot-specific movability estimates.
    
    \item \textbf{Uncertainty-aware navigation planning with probability output from VLM:} We propose an uncertainty-aware planning framework that leverages the probabilistic movability and explicitly accounts for both successful and failed interaction outcomes.
\end{itemize}

%\removed{ Fig.~\ref{fig:overview} summarizes the main components of our framework. We first review related work in Section~\ref{sec: Related Works}, followed by the VLM-based interaction feasibility estimation in Section~\ref{sec: vlm}. We then describe GSAM-based object grounding, obstacle-layered mapping, and terrain reconstruction in Section~\ref{sec: gsam and terrain reconstuction}. Finally, we present the uncertainty-aware global path planner and the Finite State Machine (FSM) in Section~\ref{sec: navigation system}. } 

%% file: Related_Works.tex
\section{Related Works}\label{sec: Related Works}

\subsection{Navigation Among Movable Obstacles}
Predictive NAMO methods often rely on restrictive assumptions about object geometry or movability supervision. For example, the work in~\cite{ellis2022navigation} considers only box-shaped obstacles, while the authors in~\cite{wang2020affordance} approximate objects using geometric primitives such as cylinders and cuboids and estimate movability through hand-designed geometric heuristics. More recently, the study of~\cite{schoch2024sight} predicts movability from DINO-based visual features but relies on game-derived data with limited object diversity and manually assigned scores that are not conditioned on robot or controller capabilities. Consequently, estimating the movability of diverse unseen objects from visual observations without privileged physical information or hand-designed heuristics remains underexplored. We address these limitations by combining VLM semantic priors with robot interaction outcomes rather than relying on manually assigned movability scores. %\Ye{[Two questions, does our approach address all the limitations of these works above? If so, we should add an ending sentence to highlight that. - Done]} 

\subsection{Rough Terrain Navigation for Humanoids}
Many NAMO approaches focus on flat indoor environments. Extending NAMO to cluttered rough terrain presents an interesting opportunity for applications such as construction sites and disaster response \cite{shamsah2025terrain}, where humanoids can leverage both dynamic locomotion and manipulation capabilities. For such rough-terrain navigation, a large body of work has studied traversability estimation for legged robots. Early approaches~\cite{wermelinger2016navigation, mccrory2023bipedal} rely on manually designed rules, while more recent methods~\cite{yoon2025state} employ neural networks to learn richer traversability representations. However, these works primarily assume static environments and do not consider movable objects, which are common in cluttered rough terrain. This limitation makes the integration of NAMO and rough-terrain navigation a particularly compelling direction.

\subsection{VLM-Based Navigation}
Recently, VLMs have demonstrated strong language-conditioned semantic understanding for navigation tasks. Various forms of semantic guidance, including social norms~\cite{song2024vlm}, path preferences~\cite{castro2025vamos}, and potential hazards~\cite{oh2025language}, have been incorporated into VLM-based navigation frameworks. These approaches generally follow one of two paradigms: end-to-end Vision-Language Navigation (VLN), or modular approaches in which VLM outputs are integrated into a classical planner.
While end-to-end VLN offers language steerability, approaches without explicit geometric representations can struggle to account for terrain traversability and object interactions in rough terrain~\cite{castro2025vamos}. Modular approaches retain classical planning while leveraging VLM semantic reasoning~\cite{oh2025language, song2024vlm}, but typically treat VLM outputs as heuristic scores without grounding them in robot capabilities or incorporating them into probabilistic planning. We instead ground VLM movability predictions in robot interaction outcomes through a residual GP, yielding embodiment-grounded probabilistic estimates for uncertainty-aware planning.

%% file: VLM.tex
\section{Object Movability Estimation with Embodiment-Grounded VLM-GP}\label{sec: vlm}
For each target object, our goal is to estimate its movability as the probability that the deployed robot and controller can successfully move the object. While a pretrained VLM provides a broad semantic prior for movability, this prior does not account for the capabilities of a specific robot. We therefore develop an embodiment-grounded VLM-GP, in which the VLM provides the semantic movability prior and a residual GP, learned from robot interaction outcomes, corrects the prior toward robot-specific movability.

\subsection{Two-Stage Object Identification and Movability Inference}\label{subsec: two stage vlm}
Inspired by the work of~\cite{oh2025language}, we decompose the inference process into two stages. First, a pretrained VLM identifies potentially movable objects from the input image and outputs their semantic labels. Given each identified object label together with the image, the embodiment-grounded VLM-GP then estimates its movability as a probability of successful interaction. This probabilistic formulation is advantageous in two respects: it provides a natural VLM prior that can be residual-corrected using robot interaction data, and it enables uncertainty-aware reasoning in the downstream planner. These aspects are described in detail in Section~\ref{subsec: vlm_finetuning} and Section~\ref{subsec: path planning}.

\subsection{Embodiment-Grounded VLM-GP for Movability Estimate}\label{subsec: vlm_finetuning}
The movability estimator takes the target object and its corresponding image as input and predicts the probability that the robot can sufficiently displace the object while maintaining stability. In principle, the mismatch between generic VLM movability priors and robot- and controller-specific capabilities could be addressed by fully adapting the model using diverse robot interaction data. However, large-scale real-world collection is impractical, and simulation still lacks sufficiently diverse, visually realistic, and physically faithful assets~\cite{cao2026physx}, particularly for objects with wheels or articulation. We therefore retain the broad semantic prior of the pretrained VLM and use the available robot interaction data to learn robot-specific corrections. Specifically, we use the VLM prediction as the prior mean and learn a residual correction with a GP. This formulation preserves the VLM prior in regimes weakly covered by robot data while introducing robot- and controller-specific corrections in regions supported by observed interaction outcomes. Accordingly, we formulate the log-odds of a successful push as:
\begin{align}
    z \;=\; z_{\text{VLM}}(I) \;+\; \Delta_{\text{GP}}(\phi), \qquad \hat{p} = \sigma(z),
\end{align}
where $\sigma(\cdot)$ denotes the logistic function and $\hat{p}$ is the estimated probability of successful interaction. The VLM semantic prior $z_{\text{VLM}}(I)$ is computed as the difference between the logits for ``Yes'' and ``No'' in response to a movability query on image $I$. The residual correction $\Delta_{\text{GP}}(\phi)$ is learned from robot interaction outcomes and conditioned on the object's physical attributes $\phi$, including mass, friction, geometry, and material class.

\begin{figure}[t]
\centerline{\includegraphicsfallback[width=1.0\linewidth]{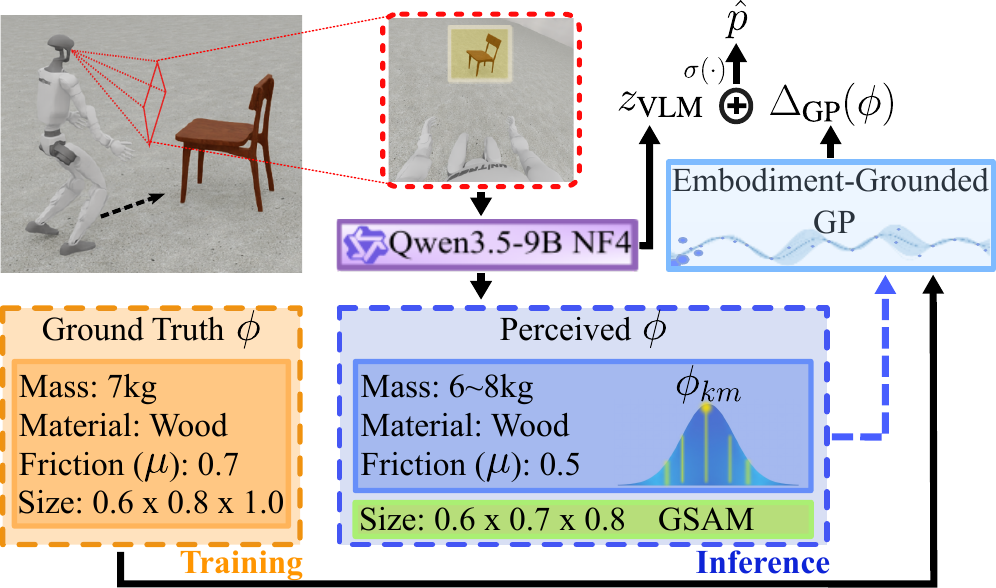}}
\caption{Overview of the proposed VLM-GP training and inference. The residual GP is trained on simulated robot interaction outcomes and ground-truth physical attributes, while perceived attributes are used at inference for embodiment-grounded movability estimation.}
\label{fig:datacollection}
  \vspace{-1.0\baselineskip}
\end{figure}

To learn the residual GP correction, we simulate the robot pushing a diverse set of objects in IsaacSim~\cite{mittal2025isaac} using the same interaction controller employed during deployment. As shown in Fig.~\ref{fig:datacollection}, the controller maintains a forward-stretched arm configuration using the whole-body controller in~\cite{zhao2026agile}, while adaptively adjusting the commanded center-of-mass (CoM) velocity to track a target pushing speed of 0.5 $\mathrm{m/s}$. An interaction is labeled successful if the robot maintains at least 10\% of the target pushing speed while displacing the object without becoming unstable, and unsuccessful otherwise. For each trial, we record the binary interaction outcome, the egocentric camera observation, the semantic name and the physical properties of the target object. We construct the simulation dataset by spawning approximately 400 furniture objects and everyday object assets derived from PhysX-3D~\cite{cao2026physx} on various ground surfaces. To improve robustness to visual and geometric variations, we randomize camera intrinsics and extrinsics, object class names, and simulation backgrounds, and additionally apply image augmentation during training. This procedure yields approximately 20,000 interaction samples. 

Using the collected interaction trials, we construct the GP training observations as follows. Trials of the same asset share mass, size, and material, so we group those whose friction falls in the same within-asset quartile into a cell $i$. Each cell has representative attributes $\phi_i$ and $s_i$ successes out of $n_i$ trials. Mass, friction, and size are simulator ground truth. The simulator does not label material, so material class is the VLM's majority reading over the asset's renders. We then fit
\begin{align}
    s_i \sim \operatorname{Binomial}\big(n_i,\, \sigma(f_i)\big), \;
    f_i = z_{\text{VLM}}(I_i) + \Delta_{\text{GP}}(\phi_i),
    \nonumber\\
    \Delta_{\text{GP}} \sim \mathcal{GP}\big(0,\, k(\cdot,\cdot)\big),
\end{align}
where $k$ is a Mat\'ern-5/2 kernel with one length scale per standardized attribute~\cite{williams2006gaussian}. By placing the VLM prediction in the GP mean function, the GP learns only the residual correction needed to align the generic semantic prior with the observed robot interaction outcomes. We approximate the posterior with the Laplace approximation~\cite{williams2006gaussian} and take its mean as $\Delta_{\text{GP}}(\phi)$. The kernel hyperparameters are estimated once by maximizing the marginal likelihood of a GP regression on the empirical log-odds of the cells and then kept fixed, which generalizes better to unseen object categories than maximizing the Laplace-approximated marginal likelihood of the Binomial model~\cite{williams2006gaussian}. As the query moves away from regions supported by the robot interaction data, the residual correction diminishes and the estimate naturally reverts toward the VLM prior.

While the GP is fitted on true simulator attributes, at test time $\phi$ must be perceived from the image. For both the VLM prior and attribute estimation, we use the same pretrained Qwen3.5-9B NF4 VLM without providing the object name. Instead, we draw a red contour along the object mask obtained from Grounded-SAM 2 (GSAM)~\cite{ren2024grounded} to visually specify the target object while reducing bias from its semantic label. To account for uncertainty in the VLM-estimated physical attributes, we query the VLM $K=5$ times for the object's mass range, friction coefficient, and material class. Object size is estimated geometrically rather than by the VLM: we lift the masked depth pixels into 3D and compute the dimensions of their bounding box. Each VLM reading is represented as a Gaussian in log space centered at its estimated value, with a standard deviation equal to twice the reported half-range width, such that the $K$ readings form a Gaussian mixture. From each reading, we draw $M=20$ samples of mass and friction, retain the corresponding material class, and combine them with the measured object size, to obtain $KM$ attribute vectors $\phi_{km}$. The final movability estimate is computed as
\begin{align}
    \hat p = \frac{1}{KM} \sum_{k=1}^{K}\sum_{m=1}^{M} \sigma\!\left(
        z_{\mathrm{VLM}}(I)
        +
        \Delta_{\mathrm{GP}}(\phi_{km})
    \right).
\end{align}
Thus, uncertainty in the perceived object attributes is marginalized rather than relying on a single VLM estimate.

%\removed{Formulating} \ziwon{Querying} movability as a probability \lu{I am a bit confused about the statement. Is our fine-tuning procedure make the VLM output a success probability or we just treat the output as probability. If it is not a probability, how did you 'formulate' it as a probability - Ziwon: We query prob in the prompt, and GP also puts correction to predict the Bernoulli outcome of the interaction} provides a practical advantage over heuristic semantic scores used in prior work~\cite{schoch2024sight, oh2025language}. Rather than requiring manual annotation of thousands of samples, training data can be generated directly from robot interaction outcomes. The resulting labels are therefore tied to a well-defined physical event and are less dependent on subjective annotator judgment. Although the success criterion itself remains a design choice, the same criterion is used during actual robot interaction, maintaining consistency between the quantity learned by the movability model and the quantity consumed by the planner. \textcolor{red}{[depending on if we need more space, this paragraph can be cut later.]} \ziwon{Agreed!}

%% file: GSAM_TerrainRecon.tex
\section{Spatial Grounding of Objects and Its Underlying Terrain Reconstruction}\label{sec: gsam and terrain reconstuction}
\begin{figure}[t]
    \centerline{\includegraphicsfallback[width=1.0\linewidth]{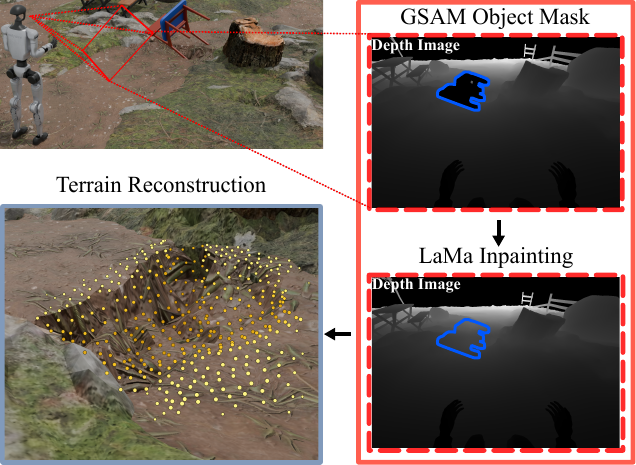}}
\caption{Spatial grounding and terrain reconstruction for a candidate object. GSAM uses the semantic label to obtain the object mask, which is applied to the aligned depth image. The masked region is then inpainted with LaMa, and both the object mask and reconstructed region are lifted into 3D to obtain the object footprint and reconstructed terrain for planning.}
\label{fig:gsam2terrainrecon}
  \vspace{-1.0\baselineskip}
\end{figure}

Section~\ref{sec: vlm} describes how the VLM-GP estimates the movability of an object from its image. To use this estimate for path planning, it must be associated with the spatial extent of the corresponding object in the bird's-eye-view (BEV) map used by the planner. We reuse the GSAM object mask obtained during movability estimation to establish this spatial correspondence. As illustrated in Fig.~\ref{fig:gsam2terrainrecon}, the masked depth pixels are lifted into a 3D point cloud and projected onto the BEV map, yielding the object footprint for planning.

While such object grounding is generally sufficient for interactive navigation on flat terrain, rough-terrain settings introduce an additional requirement. Even if an object is movable, the terrain beneath it may still be unsafe to traverse. We therefore reconstruct both the terrain geometry and its traversability beneath each candidate object. Specifically, we remove the depth pixels within the object mask and use LaMa~\cite{suvorov2022resolution} to inpaint the occluded region, providing an estimate of the terrain that would be observed in the absence of the object. The reconstructed region is then lifted into 3D and integrated into the terrain map. We subsequently apply STATE-NAV~\cite{yoon2025state} to estimate traversability, represented as the maximum speed at which the robot can safely traverse each region. If the reconstructed terrain is deemed untraversable at the target pushing speed defined in Section~\ref{subsec: vlm_finetuning}, the planner excludes the corresponding passage even when the object itself is predicted to be movable.

%% file: Path_Planning.tex
\section{Navigation System for NAMO}\label{sec: navigation system}

\begin{figure}[t]
    \centerline{\includegraphicsfallback[width=1.0\linewidth]{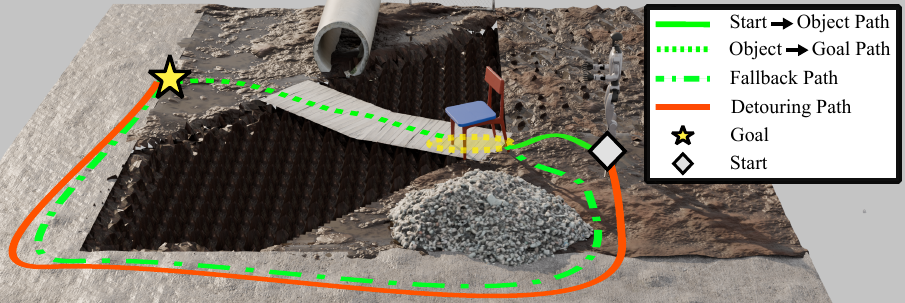}}
\caption{Illustration of uncertainty-aware path planning. The solid green path leads to the object, while the dotted and dashed green paths represent the routes following successful and failed interactions, respectively. The orange path denotes a non-interactive detour.}
\label{fig:pathplanner}
  \vspace{-1.0\baselineskip}
\end{figure}

% Emphaisze VLM output in planner context more
\subsection{Uncertainty-Aware Path Planning with VLM Output as Probability}\label{subsec: path planning}
%\Ye{[you use "Stochastic" in the title. However, it is not discussed in this section. Does the Stochastic come from the movability probability from VLM? maybe we should discuss it a bit. Or simply change the title to "Global Path Planning with VLM Output as Probability"]}

Given the estimated movability of each candidate object, the planner jointly considers travel distance, terrain traversability, and object interaction costs. Prior works incorporate VLM-derived semantic scores into classical planning~\cite{oh2025language, song2024vlm} and learned object movability scores into interactive navigation~\cite{schoch2024sight}, typically through additive costs of the form $c = c_{\mathrm{path}} + w c_{\mathrm{object}}$, where $c_{\mathrm{path}}$ is the path cost, $c_{\mathrm{object}}$ is the object-related cost, and $w$ is a tunable weight. However, such formulations do not explicitly account for uncertainty in interaction outcomes or the downstream cost of failure. We instead evaluate the expected cost of interacting with object $i$ as
\begin{align}
    c_{o_i} &= c_{\mathrm{start}\rightarrow\mathrm{obs}_i} + \hat{p}_{{o}_i} c_{\mathrm{obs}_i\rightarrow\mathrm{goal}} + (1-\hat{p}_{{o}_i}) c_{\mathrm{fallback}_i}.
\end{align}
Here, $c_{\mathrm{start}\rightarrow\mathrm{obs}_i}$ is the cost of reaching the object, $c_{\mathrm{obs}_i\rightarrow\mathrm{goal}}$ is the remaining cost after successful interaction, and $c_{\mathrm{fallback}_i}$ is the cost of retreating and reaching the goal via an alternative route after failure. As illustrated in Fig.~\ref{fig:pathplanner}, the success and failure costs are weighted by their corresponding probabilities, while non-interactive detours are evaluated directly by their path costs. The planner selects the option with minimum expected cost. For example, the planner may prefer a non-interactive detour even when an object has high predicted movability, if approaching the object incurs substantial cost or a failed interaction requires significant backtracking. All path-dependent costs are computed using terrain-aware traversal costs and A* planning~\cite{yoon2025state}, consistently accounting for traversability across successful interactions, failed interactions, and detours.

\subsection{Finite State Machine and System Integration}
When the global planner selects a route requiring object interaction, a finite state machine (FSM) switches the robot from normal path following to the pushing behavior introduced in Section~\ref{subsec: vlm_finetuning}. As the robot approaches the target, it adopts a forward-reaching posture and slowly pushes the object along the planned obstacle-passing route. During interaction, VideoSAM~\cite{ravi2024sam2} tracks the object and updates its position online. Once sufficient clearance is created, the FSM terminates the interaction and resumes path following toward the goal. For local path tracking, we employ the terrain-aware controller from~\cite{yoon2025state}.

%% file: Experimental_Results.tex
% \begin{figure}[t]
%     \centerline{\includegraphicsfallback[width=1.0\linewidth]{Figures/Results/Real_Data_Collection.pdf}}
% \caption{Real-world data collection setup. Interactable real-world objects were randomly selected and placed in front of the robot, with their initial positions shown on the left side of the figure. The robot was then commanded to push the object using the same controller and success/failure criteria as in simulation. Presented on the right side of the figure, the orange arrow indicates the direction of object movement, while the transparent robot and objects depict the pushing interaction.}
% \label{fig:RealDataCollection}
%   \vspace{-0.5\baselineskip}
% \end{figure}

\section{Experimental Results}
We organize our evaluation around five research questions (RQs). First, we evaluate the movability estimator to investigate (RQ1) how effective a pretrained VLM is for movability estimation of arbitrary objects, and how much residual GP correction using robot interaction data improves prediction and sim-to-real transfer. We then introduce the navigation benchmarks, experimental setup, and evaluation metrics used to investigate (RQ2) what new navigation capabilities the proposed framework enables, (RQ3) how residual correction using robot interaction data affects navigation performance, (RQ4) whether uncertainty-aware planning with probabilistic movability estimates improves NAMO performance on rough terrain, and (RQ5) whether the proposed navigation stack transfers to a physical humanoid in the real world.

\begin{table}[t]
    \centering
    \renewcommand{\arraystretch}{1.2}
    \caption{Object movability prediction performance on the simulation validation set (Sim) and real-world test set (Real), where Real is further divided into the Tried and Out-of-Domain (OOD) subsets. Brier scores are computed from raw predictions for Sim and from calibrated predictions for Real.}
    \label{tab:interaction_feasibility}
\begin{tabular}{l|ccc|c}
\hline
\multirow{2}{*}{Model} & \multicolumn{3}{c|}{Real} & \multirow{2}{*}{Sim} \\
\cline{2-4}
 & All & Tried & OOD & \\
\hline\hline
\multicolumn{5}{l}{\textit{AUROC} $\uparrow$} \\
\hline
\texttt{VLM-GP}          & \textbf{0.868} & \textbf{0.770} & \textbf{1.000} & 0.702 \\
\texttt{LoRA fine-tuned} & 0.817 & 0.740 & 0.960 & 0.712 \\
\texttt{Vanilla VLM}     & 0.827 & 0.681 & \textbf{1.000} & 0.662 \\
\texttt{DINOv3}          & 0.736 & 0.695 & 0.880 & \textbf{0.726} \\
\hline
\multicolumn{5}{l}{\textit{Brier} $\downarrow$} \\
\hline
\texttt{VLM-GP}          & \textbf{0.140} & 0.157 & \textbf{0.064} & 0.219 \\
\texttt{LoRA fine-tuned} & 0.148 & \textbf{0.139} & 0.187 & 0.203 \\
\texttt{Vanilla VLM}     & 0.168 & 0.180 & 0.109 & 0.276 \\
\texttt{DINOv3}          & 0.175 & 0.152 & 0.277 & \textbf{0.195} \\
\hline
\multicolumn{5}{l}{\textit{Resolution} $\uparrow$} \\
\hline
\texttt{VLM-GP}          & \textbf{0.101} & \textbf{0.045} & \textbf{0.222} & 0.035 \\
\texttt{LoRA fine-tuned} & 0.078 & 0.033 & 0.178 & 0.038 \\
\texttt{Vanilla VLM}     & 0.083 & 0.021 & \textbf{0.222} & 0.024 \\
\texttt{DINOv3}          & 0.043 & 0.021 & 0.165 & \textbf{0.045} \\
\hline
\end{tabular}%
  \vspace{-1.0\baselineskip}
\end{table}

\begin{figure*}
    \centering
    \includegraphics[width=1.8\columnwidth]{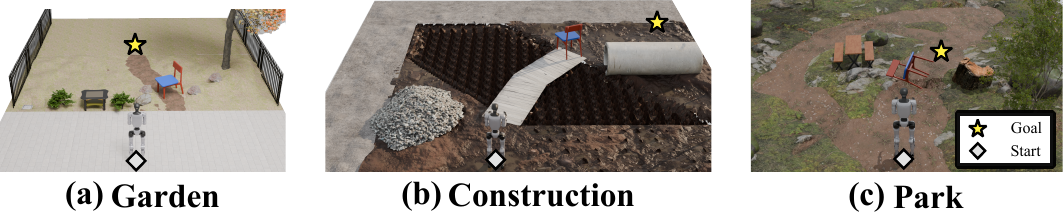}%
    \caption{Overview of the IsaacSim benchmark environments. (a) \textit{Garden}: The robot starts on the white tiles and reaches the goal through the left opening or right dirt path, both blocked by movable obstacles. (b) \textit{Construction}: The wooden plank bridge offers a shorter route across uneven terrain than the detour along the concrete boundary. (c) \textit{Park}: The shorter route is blocked by an object on highly uneven terrain, where pushing it may cause the robot to trip.
    }
    \label{fig:sim_env}
  \vspace{-1.0\baselineskip}
\end{figure*}

\subsection{VLM-GP Movability Estimation and Sim-to-Real Transfer}

\subsubsection{Movability Estimation Setup}
To address RQ1, we first analyze the accuracy and sim-to-real transfer of the proposed movability estimator. For the simulation data collected in Section~\ref{subsec: vlm_finetuning}, we split the interaction samples into training and validation sets. To evaluate sim-to-real generalization, we additionally collect a real-world dataset following the same interaction procedure. %As shown in Fig.~\ref{fig:RealDataCollection},
We place diverse objects on multiple floor surfaces and evaluate their movability using the same pushing behavior as in simulation. In total, we physically test 73 objects, including five wheeled objects absent from the simulation training data. We additionally collect images of 10 clearly immovable objects, such as a public bench and a heavy refrigerator, without conducting physical interactions. We define three evaluation sets: \textit{Real-Tried}, comprising the 68 physically tested non-wheeled objects; \textit{Real-OOD}, comprising the five wheeled objects and 10 additional immovable objects; and \textit{Real-All}, containing all 83 objects. Thus, Real-Tried and Real-OOD are disjoint subsets of Real-All.

We compare the proposed VLM-GP against three alternative movability estimators. The first is the pretrained Qwen3.5-9B NF4 VLM without residual correction. The second is a low-rank adaptation (LoRA) fine-tuned version of the same VLM, trained on the simulated interaction data using binary cross-entropy over the ``Yes'' and ``No'' outputs. The third is a DINO-based predictor inspired by~\cite{schoch2024sight}, using DINOv3~\cite{simeoni2025dinov3} to extract visual features from the masked object region, followed by a multilayer perceptron (MLP) trained on the same simulated interaction data. We evaluate all methods on the simulation validation set and the real-world test set using the area under the receiver operating characteristic curve (AUROC), Brier score, and the resolution component of the Brier score decomposition. Higher AUROC indicates better discrimination, while lower Brier score indicates better probabilistic accuracy. However, the Brier score can favor conservative predictions near the empirical success rate even when they are weakly discriminative. We therefore also report resolution, where higher values indicate greater separation among samples with different empirical success rates and thus more informative probabilities for downstream planning. To account for the domain shift in probability estimates, we reserve 20\% of the real-world data for post-hoc calibration~\cite{guo2017calibration} and report the real-world probability metrics on the remaining held-out data.

\subsubsection{Performance Analysis}
As shown in Table~\ref{tab:interaction_feasibility}, the DINO-based predictor and the LoRA fine-tuned VLM slightly outperform the VLM-GP and pretrained VLM on the simulation validation set. However, their performance degrades substantially on the real-world data. Although the LoRA fine-tuned VLM and DINO-based predictor achieve lower Brier scores than VLM-GP on Real-Tried, both learned baselines exhibit substantially lower resolution, indicating that their favorable Brier scores partly arise from conservative predictions with limited discrimination. This suggests that directly adapting high-dimensional visual representations to simulated interaction data can lead to poorer transfer to real-world observations. By comparison, VLM-GP retains the pretrained VLM prior and learns only a residual correction in a low-dimensional physical attribute space, which is less sensitive to visual domain shift.

The effect of residual correction is particularly evident on Real-Tried, where VLM-GP shows a larger improvement over the pretrained VLM than on Real-All or Real-OOD. This indicates that the GP provides useful robot-specific corrections when the test objects are supported by the interaction data. The LoRA fine-tuned and DINO-based predictors degrade more strongly on Real-OOD, whereas VLM-GP is less affected because its residual correction diminishes away from the training distribution and the prediction reverts toward the pretrained VLM prior.

Finally, the raw pretrained VLM exhibits particularly poor Brier performance on the uncalibrated simulation set despite providing useful semantic movability predictions. Unlike the raw VLM, the learned estimators are trained directly on binary robot interaction outcomes, providing supervision on the probability scale of interaction success. Representing movability as an event probability rather than an unscaled semantic score also allows the prediction to be refined or calibrated using interaction outcomes directly obtainable from robot execution. Such embodiment-grounded probabilistic processing is important for uncertainty-aware planning, since directly treating raw VLM outputs as interaction-success probabilities can produce poorly scaled estimates and distort downstream planning decisions.

\subsection{Experimental Setup for Planner Testing}
We deploy various planners on a Unitree G1 robot equipped with an RGB-D camera. Planners must generate a path and motion plan that guide the robot safely through obstacle-cluttered rough terrain while minimizing instability, clearing objects when necessary, and reaching the goal efficiently.

% TODO: Keep the columns grouped by Garden, Construction, and Park, with SR, Ins., $t$, and Push \% as the four subcolumns for each environment.
% TODO: Keep one row per planner benchmark, with Pro-VanillaVLM and Pro-DINO in separate rows. Map SR to succ, Ins. to instab max, $t$ to nav_t[s], and Push \% to thru_succ/thru_plan when filling the table. Display Push \% using both source percentages in that order (e.g., 30\% / 60\%) and bold the entry with the highest thru_succ-to-thru_plan ratio.
% TODO: VLMScore corresponds to the nounc_w* results. Select one best-performing nounc setting independently for each environment, use all metrics from that selected setting, and show the selected values under VLMScore as (w = {Garden}, {Construction}, {Park}).
% TODO: After replacing the placeholders with experimental results, bold the best planner value in each environment and metric. Do not fill the table values until explicitly requested.

\begin{table*}[t]
    \centering
    \renewcommand{\arraystretch}{1.5}
    \caption{Navigation performance in \textit{Garden}, \textit{Construction}, and \textit{Park}. N-SR denotes goal-reaching success with body tilt at most $45^{\circ}$. Mean $T$ and $T$-CVaR$_{0.9}$ are computed over successful runs, with the latter averaging the slowest 10\%. I-SR denotes the success rate of attempted interactions. For Pro-NoUnc, $w$ is tuned independently for each environment based on N-SR. Bold indicates the best value in each column, and parentheses show relative degradation computed from unrounded values. $^\dagger$ STATE-NAV cannot find a path because obstacles block all routes. A dash indicates unavailable metrics, including I-SR when no interaction is attempted.}
    \label{tab:navigation results}
    \resizebox{\textwidth}{!}{%
\begin{tabular}{l|cccc|cccc|cccc}
\hline
 & \multicolumn{4}{c|}{Garden} & \multicolumn{4}{c|}{Construction} & \multicolumn{4}{c}{Park} \\
\cline{2-13}
 Planner & N-SR & $T$ (Mean, s) & $T$ (CVaR$_{0.9}$, s) & I-SR & N-SR & $T$ (Mean, s) & $T$ (CVaR$_{0.9}$, s) & I-SR & N-SR & $T$ (Mean, s) & $T$ (CVaR$_{0.9}$, s) & I-SR \\
\hline\hline
Proposed       & \textbf{95.5\%} & 14.6 (+16\%) & 20.0 (+13\%) & 91.1\% & 97.7\% & 20.3 (+5.7\%) & \textbf{24.9 (+0\%)} & \textbf{100\%} & \textbf{100\%} & \textbf{17.8 (+0\%)} & \textbf{19.0 (+0\%)} & - \\
Pro-NoRecon    & \textbf{95.5\%} & 14.4 (+14\%) & 18.7 (+5.6\%) & 93.5\% & 93.2\% & 21.2 (+10\%) & 33.0 (+33\%) & 96.0\% & 86.4\% & 18.9 (+6.5\%) & 28.6 (+51\%) & 0.0\% \\
Pro-VanillaVLM & 93.2\% & 16.5 (+31\%) & 30.4 (+72\%) & 84.0\% & 97.7\% & 22.4 (+17\%) & 25.2 (+1.4\%) & 90.0\% & 93.2\% & 17.8 (+0.3\%) & 19.7 (+3.7\%) & - \\
Pro-NoUnc (Best $w$) & \textbf{95.5\%} & 14.4 (+14\%) & 21.0 (+19\%) & \textbf{100\%} & 97.7\% & 20.0 (+4.0\%) & 32.6 (+31\%) & 84.6\% & \textbf{100\%} & 18.0 (+1.1\%) & 20.9 (+10\%) & 0.0\% \\
STATE-NAV      & 0\%$^\dagger$ & - & - & - & \textbf{100\%} & 24.6 (+28\%) & 25.0 (+0.5\%) & - & 97.7\% & \textbf{17.8 (+0\%)} & 19.3 (+1.7\%) & - \\
VAMOS          & 52.3\% & \textbf{12.6 (+0\%)} & \textbf{17.7 (+0\%)} & 75.6\% & 63.6\% & \textbf{19.2 (+0\%)} & 42.2 (+70\%) & 51.9\% & 11.4\% & 21.3 (+20\%) & 52.8 (+180\%) & \textbf{42.9\%} \\
\hline
\end{tabular}%
}
\vspace{-0.2in}
\end{table*}

\subsubsection{Planner Benchmarks}
We evaluate six planners to isolate the contributions of the proposed framework:

\begin{itemize}
    \item \textbf{Proposed}: Uses VLM-GP movability estimates for uncertainty-aware planning while reconstructing terrain traversability beneath objects.

    \item \textbf{Pro-NoRecon}: Assumes the terrain beneath objects is traversable without reconstruction. All other components remain identical to Proposed.

    \item \textbf{STATE-NAV~\cite{yoon2025state}}: A non-interactive rough-terrain planner that treats objects as terrain obstacles while sharing the traversability estimator and underlying planner with Proposed.

    \item \textbf{Pro-VanillaVLM}: Replaces VLM-GP with the uncalibrated pretrained VLM evaluated in Table~\ref{tab:interaction_feasibility}, leaving all other components unchanged.

    \item \textbf{VAMOS~\cite{castro2025vamos}}: An end-to-end VLN method that generates candidate paths from language instructions, from which we select the one with the lowest traversability cost. Among the tested prompts, the best-performing is ``$\{\rm object\}$ is $\{\rm probability\}$ likely to be clearable. Clear it and continue, or go around it.'', where both variables are provided by the same VLM-GP used in Proposed for a fair comparison.

    \item \textbf{Pro-NoUnc (similar to~\cite{oh2025language, schoch2024sight})}: Treats VLM-GP movability estimates as scalar scores in an additive planning objective, $c = c_{\rm path} + w \cdot (1-\hat{p}_{o_i})$, where $w$ is a tunable weight, following the score-based planning structure of prior work.
\end{itemize}

\subsubsection{Simulation Benchmark}
Existing navigation benchmarks typically represent objects as static meshes and lack outdoor rough terrain, making them unsuitable for NAMO evaluation. We therefore construct three IsaacSim environments, shown in Fig.~\ref{fig:sim_env}. To prevent asset overlap with the simulation training data, we reserve 11 randomly selected objects exclusively for evaluation. Each planner is evaluated over 44 trials per environment using four random seeds. In each trial, the robot starts on flat terrain and is given 11~s to rotate, scan the environment, and initialize its plan toward a predefined goal. All experiments run on a desktop equipped with an Intel Core Ultra 9 285 CPU, 64~GB RAM, and an NVIDIA GeForce RTX 5090 GPU with 32~GB VRAM.

\subsubsection{Real-World Validation}\label{sec: hardware results}
For real-world validation, we deploy the proposed framework on a Unitree G1 equipped with an NVIDIA Jetson AGX Orin (64~GB) and a ZED X camera, supported by an external laptop with an RTX 5090 Laptop GPU. The robot's Livox LiDAR provides point clouds, while the ZED X supplies RGB and depth images. FAST-LIO~\cite{xu2022fast} performs localization on the G1's built-in Jetson Orin NX, while all other modules run on the AGX, except for VLM and GSAM inference, which are executed on the external laptop and transmitted to the AGX over the network. We retain the simulation settings except for the computational budgets of STATE-NAV's A* and model predictive path integral (MPPI) modules.

\begin{figure}[H]
    \centering
    \includegraphicsfallback[width=0.9\columnwidth]{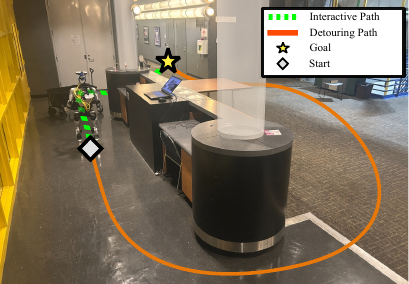}%
    \caption{Illustration of the real-world testing environment with path plan candidates. The green dashed line represents a route to the goal that involves interaction with a black wagon, while the orange line represents an alternative detouring route.}
    \label{fig:indoor_exp}
  \vspace{-1.0\baselineskip}
\end{figure}

\begin{figure}
    \centering
    \includegraphicsfallback[width=0.9\columnwidth]{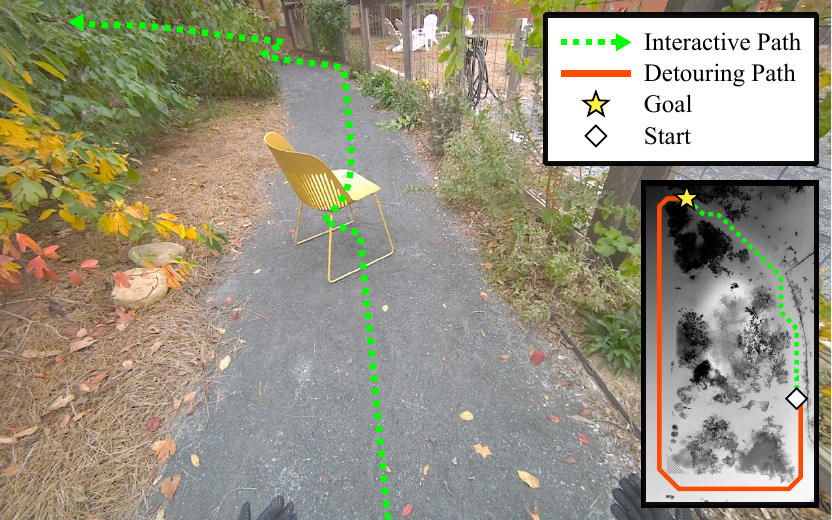}%
    \caption{Outdoor testing environment (22~m $\times$ 41~m) with a shortcut path blocked by a chair. The green path indicates the route selected by the robot, which clears the chair to take the shortcut. The inset on the lower right shows a top-down view of the entire test site, including the selected shortcut path and the alternative orange detour path.}
    \label{fig:outdoor}
  \vspace{-1.0\baselineskip}
\end{figure}

\begin{figure}
    \centering
    \includegraphicsfallback[width=1.0\columnwidth]{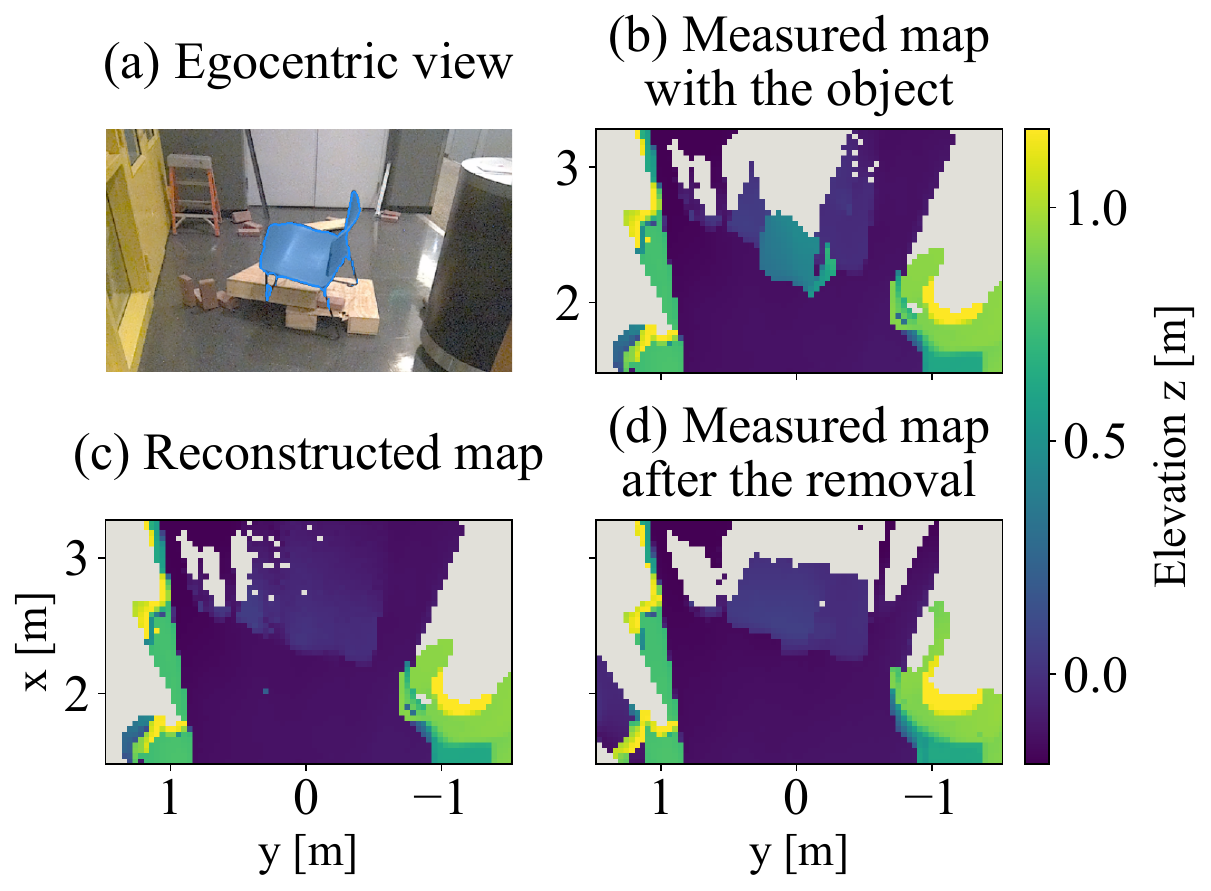}%
    \caption{Real-world terrain reconstruction on rough terrain. (a) Egocentric view of a chair resting on an artificial rough-terrain setup constructed in the indoor environment shown in Fig.~\ref{fig:indoor_exp}. (b) BEV elevation map measured with the chair present. (c) Terrain map reconstructed using the method in Section~\ref{sec: gsam and terrain reconstuction}. (d) BEV elevation map measured after the chair is removed.}
    \label{fig:terrain_reconstruction}
  \vspace{-1.0\baselineskip}
\end{figure}

As illustrated in Fig.~\ref{fig:indoor_exp}, we evaluate the framework in an indoor environment where obstacles are added to or removed from a previously mapped scene, requiring the navigation system to adapt to the changes. We test four objects available at the site, including a wagon used to transport the robot, tables, and a chair in the lobby. Each planner is evaluated in two trials per object, totaling eight trials. Given the limited sample size, we replace $T$-CVaR$_{0.9}$ with mean navigation distance, as reliable tail-risk estimation would require more trials. To evaluate the contribution of terrain reconstruction through the Pro-NoRecon ablation, we additionally construct rough terrain at the test site, as shown in Fig.~\ref{fig:terrain_reconstruction}. Finally, we demonstrate the feasibility of the proposed framework in a natural outdoor environment, as shown in Fig.~\ref{fig:outdoor}.

\subsubsection{Performance Metrics}
We evaluate navigation success rate (N-SR), mean navigation time ($T$), conditional value at risk ($T$-CVaR$_{0.9}$), and interaction success rate (I-SR). A run succeeds if the robot reaches the goal without its body tilt exceeding $45^{\circ}$. N-SR is computed over all runs, while mean $T$ and $T$-CVaR$_{0.9}$ use successful runs only. The latter measures tail performance as the average navigation time among the slowest 10\% of successful runs. I-SR is the fraction of successful obstacle-pushing interactions among all attempts. Since all planners share the same interaction controller, a higher I-SR indicates more reliable selection of objects for interaction.

\subsection{Planner Testing Results}

\textbf{\textit{RQ2: What new navigation capability does the proposed framework enable?}}
We examine the navigation capabilities enabled by the proposed framework across the planner benchmarks in Table~\ref{tab:navigation results}. In \textit{Garden}, the non-interactive STATE-NAV planner cannot generate a path because movable objects block every route to the goal. In \textit{Construction}, STATE-NAV achieves a 100\% N-SR but a higher mean $T$ because it cannot exploit the shorter interactive route, whereas successful interactive runs complete the task in approximately 15~s. In \textit{Park}, Pro-NoRecon exhibits a notably lower N-SR and higher $T$-CVaR$_{0.9}$ because it more frequently selects obstacle-passing routes without accounting for the traversability of the terrain beneath the object. This leads to falls or backtracking when the revealed terrain cannot be safely traversed, demonstrating the importance of reasoning about the terrain beneath movable objects. By jointly reasoning about object movability and terrain traversability, the proposed framework enables reliable navigation through cluttered rough terrain.

\textit{\textbf{RQ3: Does embodiment-grounded VLM-GP improve navigation performance?}}
%\subsubsection{Does embodiment-grounded VLM-GP improve navigation performance?}
To assess how movability estimation affects navigation, we compare the proposed VLM-GP with the vanilla VLM beyond the prediction-level results in Table~\ref{tab:interaction_feasibility}. The vanilla VLM exhibits a substantially higher Brier score, and Pro-VanillaVLM correspondingly achieves a lower N-SR in Table~\ref{tab:navigation results}. More importantly, I-SR degrades notably in \textit{Garden} and \textit{Construction}, indicating that inaccurate movability estimates can lead the planner to select difficult-to-move objects or avoid favorable interactions. These errors also worsen tail performance, with Pro-VanillaVLM exhibiting a higher $T$-CVaR$_{0.9}$ in all environments, particularly in \textit{Garden}. Overall, more accurate embodiment-grounded movability estimates enable better interaction decisions and improve both navigation reliability and tail performance.

\begin{figure}
    \centering
    \includegraphicsfallback[width=1.0\columnwidth]{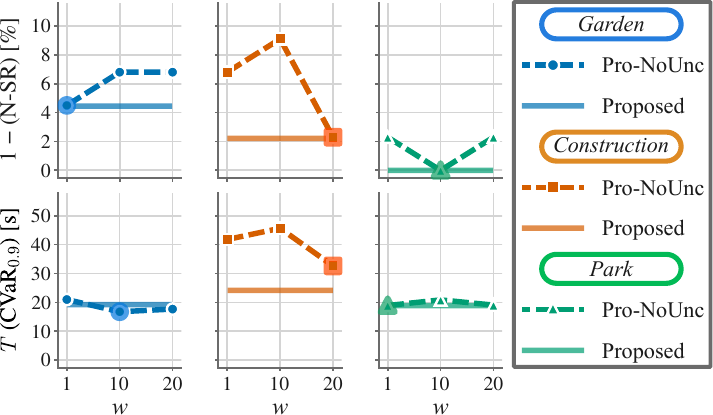}%
    \caption{Sensitivity to $w$ across three environments. The top and bottom rows show navigation failure rate ($1-\mathrm{N\text{-}SR}$) and $T$-CVaR$_{0.9}$, respectively. Lower values indicate better performance for both metrics. Dashed lines represent Pro-NoUnc with varying $w$, while solid lines represent the proposed method without weight tuning. }
    \label{fig:w_sensitivity}
  \vspace{-1.0\baselineskip}
\end{figure}

\textbf{\textit{RQ4: Is uncertainty-aware planning with probabilistic VLM outputs effective for NAMO on rough terrain?}}
Finally, we compare the proposed planner with VAMOS and Pro-NoUnc to examine planning formulations that jointly reason about terrain traversal, object interaction, and alternative routes. As shown in Table~\ref{tab:navigation results}, VAMOS performs notably worse than the other methods and exhibits substantially higher $T$-CVaR$_{0.9}$ in \textit{Construction} and \textit{Park}. This degradation appears to reflect limitations in the global spatial reasoning of its end-to-end VLN formulation. Without an explicit spatial map over its observation history, VAMOS often fails to identify globally preferable detours and frequently selects object interaction when an object is mentioned in the prompt. Moreover, in \textit{Garden}, when both candidate objects are visible in the same field of view, VAMOS consistently moves toward the object closer to the image center despite differences in their estimated movability. These observations indicate difficulty in comparing competing interaction and detour options and highlight the benefit of explicit spatial representations with persistent memory for long-horizon navigation.

Pro-NoUnc exhibits a different limitation. For Table~\ref{tab:navigation results}, we report its best-performing configuration from an environment-specific sweep over the weight $w$. Although this tuning yields N-SR and mean $T$ comparable to the proposed method, its I-SR and $T$-CVaR$_{0.9}$ are notably worse in \textit{Construction} despite using the same movability estimator, indicating that the difference arises from the planning formulation. Pro-NoUnc combines path-traversal and object-related costs through an additive weighted sum without explicitly accounting for the consequences of successful and failed interactions, so a favorable weighted score can still lead to a costly fallback after interaction failure. Moreover, the best weights differ substantially across environments, with $w=1$, $20$, and $10$ for \textit{Garden}, \textit{Construction}, and \textit{Park}, respectively, and Fig.~\ref{fig:w_sensitivity} shows strong performance sensitivity to $w$ within each environment. This sensitivity arises from directly combining quantities with different meanings and scales, making the appropriate weight dependent on the environment and route geometry. By contrast, the proposed formulation treats movability as the probability of interaction success and uses it to combine the navigation costs of the corresponding success and failure outcomes within a common cost domain. Together, these results indicate that explicit spatial memory and uncertainty-aware probabilistic planning provide a more robust formulation for humanoid NAMO on rough terrain.

\begin{table}[t]
    \centering
    \renewcommand{\arraystretch}{1.5}
    \caption{Real-world navigation performance. N-SR denotes goal-reaching success without safety intervention by a human operator. Mean navigation time ($T$) and distance are computed over successful runs. Bold indicates the best value in each column, and parentheses show relative degradation from the best value. $^\dagger$ VAMOS and Pro-NoRecon are evaluated only on the chair and wagon. All attempted trials require early termination due to imminent instability, yielding 0\% N-SR on this limited subset. Further testing is discontinued for hardware safety. A dash indicates unavailable metrics or no interaction attempts.}
    \label{tab:real_world}
    \resizebox{\columnwidth}{!}{%
    \begin{tabular}{l|ccc}
    \hline
    Planner & N-SR & $T$ (Mean, s) & Distance (Mean, m) \\
    \hline\hline
    Proposed
        & \textbf{100\%} & \textbf{23.2 (+0\%)} & \textbf{10.3 (+0\%)} \\
    Pro-VanillaVLM
        & \textbf{100\%} & 29.2 (+26\%) & 16.7 (+62\%)  \\
    STATE-NAV
        & \textbf{100\%} & 27.6 (+19\%) & 16.8 (+61\%) \\
    VAMOS
        & 0\%$^\dagger$ & - & - \\
    Pro-NoRecon
        & 0\%$^\dagger$ & - & - \\
    \hline
    \end{tabular}%
    }
    \vspace{-0.2in}
\end{table}

\textbf{\textit{RQ5: Does the proposed navigation stack transfer to the real world?}}
The real-world results in Table~\ref{tab:real_world} evaluate the feasibility of deploying the proposed framework on a physical humanoid. Given the limited number of trials, we focus on Proposed, Pro-VanillaVLM, and STATE-NAV to illustrate how movability estimation affects navigation decisions. For the three movable objects, the vanilla VLM predicts an average movability of 0.33, whereas the proposed embodiment-grounded VLM-GP estimates 0.90. Consequently, Proposed favors interaction routes, while Pro-VanillaVLM and the non-interactive STATE-NAV favor detours. Although all three planners achieve 100\% N-SR, Proposed reduces both navigation time and distance. We also test VAMOS in two trials, both of which require manual termination due to imminent instability, prompting us to discontinue further testing for hardware safety.

To assess terrain reconstruction, we additionally introduce artificial rough terrain consisting of wooden boards and bricks beneath the chair and the wagon. As shown in Fig.~\ref{fig:terrain_reconstruction}, the proposed method reconstructs the underlying terrain with an average height RMSE of 5.0~cm across the chair and wagon tests, although some sharp terrain features are smoothed. This reconstruction enables Proposed to identify the underlying terrain as hazardous and select a detour, whereas Pro-NoRecon attempts to traverse the same region and requires early termination due to imminent instability. A video of the Pro-NoRecon failure is available on our project website.

Finally, we demonstrate the framework in a natural outdoor environment, as shown in Fig.~\ref{fig:outdoor}, where the robot selects a shortcut by clearing a chair blocking the path. Together, these preliminary experiments demonstrate the feasibility of the integrated framework on a physical humanoid and suggest that embodiment-grounded movability estimation and terrain-aware planning improve navigation efficiency and robustness.

%% file: Conclusion_and_Future_Works.tex
\section{Conclusion and Future Works}
In this paper, we present a framework that extends humanoid navigation beyond obstacle avoidance to physical interaction on rough terrain. Our embodiment-grounded VLM-GP combines pretrained semantic priors with robot interaction outcomes to produce embodiment-specific probabilistic movability estimates. By integrating these estimates with object grounding, underlying terrain reconstruction, and uncertainty-aware planning, the framework evaluates interaction opportunities while accounting for potential failure costs. Extensive simulation experiments and real-world validation demonstrate the effectiveness and feasibility of the proposed framework.

Nevertheless, the current framework relies on a relatively simple pushing strategy without explicitly optimizing object displacement for subsequent navigation. Scalability to densely cluttered environments also remains challenging, particularly when distinguishing multiple object instances and repeatedly evaluating their movability using VLMs. Future work will investigate more sophisticated interaction strategies and scalable perception and planning to further advance long-horizon humanoid NAMO in complex environments.

\section*{Acknowledgment}
LLMs, including ChatGPT and Claude Code, are used in polishing the writing of this paper. In particular, spelling and grammar structure are aided by LLMs. Additionally, Claude Code is used to clean and polish our code base. This aids in formatting and removing deprecated code.